\documentclass[12pt]{article}
\usepackage{amsmath}
\usepackage{amssymb}   
\usepackage{enumitem}
\usepackage{graphicx}
\usepackage{amssymb}
\usepackage[section]{placeins}
\usepackage{fancybox}
\usepackage{rotating}
\DeclareMathAlphabet{\mathpzc}{OT1}{pzc}{m}{it}

\newtheorem{theorem}{Theorem}

\newcommand{\inda}{\phantom{1}\hspace{3mm}}
\newcommand{\indb}{\phantom{1}\hspace{8.5mm}}
\newcommand{\indc}{\phantom{1}\hspace{12mm}}
\newcommand{\indd}{\phantom{1}\hspace{16mm}}
\newcommand{\inde}{\phantom{1}\hspace{20mm}}

\newcommand{\by}{\boldsymbol{y}}
\newcommand{\bt}{\boldsymbol{t}}

\begin{document}

%

\title{Notes on hierarchical ensemble methods for DAG-structured taxonomies}

\author{{\em Giorgio Valentini}\\ 
        DI - Dipartimento di Informatica \\
	Universit\`a degli Studi di Milano \\
	e-mail: valentini@di.unimi.it}
	
\date{}
\maketitle

\begin{abstract}
Several real problems ranging from text classification to computational biology are characterized by hierarchical multi-label classification tasks. 
Most of the methods presented in literature focused on tree-structured taxonomies, but only few on taxonomies structured according to  a Directed Acyclic Graph (DAG).
In this contribution novel classification ensemble algorithms for DAG-structured taxonomies are introduced. In particular Hierarchical Top-Down ({\em HTD-DAG}) and True Path Rule ({\em TPR-DAG}) for DAGs are presented and discussed.
\end{abstract}

\section{Introduction}
Hierarchical classification problems are characterized by taxonomies structured according to a pre-defined hierarchy. 
Examples in the context of the gene or protein function  prediction include trees or  directed acyclic graphs~\cite{GOcons00}, where functional classes are connected according to a tree (FunCat, Functional Categories~\cite{Ruepp04}) or a DAG (GO, Gene Ontology~\cite{GOcons00}). 

Extensive experimental studies showed that flat prediction, i.e. predictions  for each class made independently of the other classes, introduce significant inconsistencies in the classification, due to the violation of the {\em true path rule}, that governs the hierarchical relationships between classes~\cite{Obozinski08,Vale12a}.
According to this rule, positive predictions for a given term must be transferred to its ``ancestor'' terms and negative predictions to its descendants.  

In their more general form hierarchical ensemble methods adopt a two-steps learning strategy~\cite{Jiang08,Vale10b,Cerri10b,Schietgat10}:
\begin{enumerate}
\item In the first step each base learner separately or interacting with connected base learners learns the protein functional category on a per-term basis. In most cases this yields a set of independent classification problems, where each base learning machine is trained to learn a specific functional term, independently of the other base learners.
\item In the second step the predictions provided by the trained classifiers are combined by considering the hierarchical relationships between the base classifiers modeled according to the hierarchy of the functional classes.
\end{enumerate}
Usually they significantly improve prediction performances with respect to ``flat'' prediction methods~\cite{Guan08,Vale11a,Reddy12}.

Hierarchical classification and in particular ensemble methods for hierarchical classification have been applied in several domains ranging from protein function prediction~\cite{Eisner05,Blockeel06,Astikainen08}, to text categorization~\cite{Koller97,Chakrabarti98,ZhangML06}, to music genre classification~\cite{Burred03,DeCoro07,Tsoumakas08a}, hierarchical image classification~\cite{Binder09,Barutcuoglu06a} and video annotation~\cite{Dimou09}, and automatic classification of World Wide Web documents~\cite{Punera08,Ceci07}.
A general review on hierarchical classification methods and their applications in different domains is provided in~\cite{Silla11}, while a review on hierarchical ensemble methods in the context of computational biology is provided in~\cite{Vale14c}.

Most of the proposed ensemble methods have been proposed for tree-structured taxonomies~\cite{Cerri10,Cerri11,Vale11a,Hernandez13,Vale10b,CB06,Vale12a,Chen12b}, and only a few for DAG-structured taxonomies~\cite{Obozinski08,Guan08,Schietgat10}.

In this contribution we propose and discuss novel hierarchical ensemble methods for DAGs, in particular the Hierarchical Top-Down ({\em HTD-DAG}), the True Path Rule ({\em TPR-DAG}) algorithms and some their variants specifically designed for DAG-structured taxonomies.

\section{Basic notation and definitions}
\label{sec:basic}
Let $G = <V,E>$ a Directed Acyclic Graph (DAG) with vertices $V = \{1, 2, \ldots, |V| \}$ and edges $e=(i,j) \in E, i,j \in V$. $G$ represents a taxonomy structured as a DAG, whose nodes $i \in V$ represent classes of the taxonomy and a directed edge $(i,j) \in E$ the hierarchical relationships between $i$ and $j$: $i$ is the parent class and $j$ is the child class. 
We assume that does exist a unique root node $root(G)$ of the DAG; if there are multiple roots we can easily add a unique root node by simply adding an edge between the added root and the original multiple roots.

The set of children of a node $i$ is denoted $child(i)$, the set of its parents  $par(i)$, the set of its ancestors $anc(i)$ and the set of its descendants $desc(i)$.

A ``flat continuous'' classifier $f: X \rightarrow \mathbb{Y}$ provides a score $\hat{\by} \in \mathbb{Y}=[0,1]^{|V|}$ for a given example $x \in X$. In other words a flat classifier provides a score $\hat{y}_i \in [0,1]$ for each node/class $i \in V$ of the DAG $G$:
\begin{equation}
\hat{\by} = < \hat{y}_1, \hat{y}_2, \ldots, \hat{y}_{|V|}>
\end{equation} 
It is easy to see that a ``flat discrete'' binary classifier is a special case of $f$, where  $\hat{y}_i \in \{0,1\}$. 
In this case the classifier simply assigns  ($\hat{y}_i = 1$) or not ($\hat{y}_i = 0$) an example $x$ to class $i$. 
In the more general case the flat ``continuous'' classifier provides  scores $\hat{y}_i \in [0,1]$ that can be interpreted as the likelihood or probability of belonging to a given class $i$.

In the case of a ``flat discrete'' classifier the labels $\hat{\by}$ directly identify the set
$S \subset V$ of the predicted classes:
\begin{equation} 
S = \{ i | i \in V \wedge \hat{y}_i = 1 \}
\end{equation} 

We say that $S$ is a valid ``discrete'' labeling if the following property ({\em true path rule}) holds:
\begin{equation} 
S \; {\rm is \; valid}  \iff S = \{ i | i \in V \wedge i \in S \wedge j \in par(i) \Rightarrow j \in S \}
\label{eq:discr-true-path-rule}
\end{equation} 
In the more general case of continuous scores, we say that the multi-label scoring $\by$ is valid if it respects the {\em true path rule}:
\begin{equation} 
\by \; {\rm is \; valid}  \iff \forall i \in V, i \in par(j) \Rightarrow y_i \geq y_j 
\label{eq:cont-true-path-rule}
\end{equation}
It is easy to see that (\ref{eq:discr-true-path-rule}) is a special case of (\ref{eq:cont-true-path-rule}).

We say that a classifier satisfies the true path rule if for any example $x$, for respectively a flat discrete or a  continuous classifier property (\ref{eq:discr-true-path-rule}) or (\ref{eq:cont-true-path-rule}) holds.

In real cases it is very unlikely that a flat discrete or continuous classifier satisfies the true path rule, since by definition the predictions are performed without considering the hierarchy of the classes. 
Nevertheless by adding a further label/score modification step, i.e. by taking into account the hierarchy of the classes, we can modify the labeling or the scores of the flat classifiers to obtain a hierarchical classifier that obeys the true path rule. 
In other words we can provide a function $h(f(x)): X \rightarrow \mathbb{Y}$ such that $\forall x \in X$, given that $f(x) = \hat{\by}$ and $h(f(x)) = \bar{\by}$, we have for all $(x, \bar{\by})$:
\begin{equation} 
 \forall i \in V, i \in par(j) \Rightarrow \bar{y}_i \geq \bar{y}_j 
\label{eq:vlid-hier-corr}
\end{equation}
In this contribution we propose several hierarchical algorithms that implement such function $h$.

\section{Hierarchical top-down algorithm for DAGs (HTD-DAG)}
\label{sec:HTD}
The simplest hierarchical algorithm for DAG is the Hierarchical top-down algorithm {\em (HTD)}.
It adopts this simple following rule by per-level visiting the nodes from top to bottom:
\begin{equation}
\bar{y}_i = \left\{
                 \begin{array}{lll}
				   \hat{y}_i  & {\rm if} \quad i \in root(G) \\
                   \min_{j \in par(i)} \bar{y}_j & {\rm if} \quad \min_{j \in par(i)} \bar{y}_j < \hat{y}_i \\
                   \hat{y}_i & {\rm otherwise}
                 \end{array}
                \right.
\label{eq:HTD} 
\end{equation}
Note that the level must correspond to the maximum distance of the node from the root.

\begin{theorem} Given a DAG $G = <V,E>$, a set of flat predictions $\hat{\by} = < \hat{y}_1, \hat{y}_2, \ldots, \hat{y}_{|V|}>$ for each class associated to each node $i \in \{1, \ldots, |V| \}$,  the {\em HTD-DAG} algorithm assures that $\forall i \in V, \; i \in anc(j) \Rightarrow \bar{y}_i \geq \hat{y}_i$.
\end{theorem}
The proof by induction is straightforward and omitted for brevity.

\begin{figure}[!hb]
\begin{center}
\cornersize*{8pt}
\ovalbox{
\begin{minipage}{0.90\textwidth}
\caption{\bf{Hierarchical Top-Down algorithm for DAGs (HTD-DAG)}}
\label{fig-HTD}

\noindent
{\tt Input}:\\
- $G = <V,E>$\\
- $V = \{1, 2, \ldots, |V| \}$\\
- $\hat{\by} = < \hat{y}_1, \hat{y}_2, \ldots, \hat{y}_{|V|}>, \quad  \hat{y}_i \in [0,1]$\\
\noindent
{\tt begin algorithm}\\
01:  \inda   A. Compute $\forall i \in V$ the max distance from $root(G)$:\\
02:    \indb $E' := \{e' | e \in E, \, e' = - e \}$\\
03:    \indb $G' := < V, E' >$\\
04:    \indb $dist := $ Bellman.Ford$(G', root(G')$ \\
05:  \inda   B. Per-level top-down visit of $G$:\\
06:    \indb $\bar{y}_{root(G)} := \hat{y}_{root(G)}$\\
07:    \indb {\tt for each} $d$ {\tt from} $1$ {\tt to} $\max(dist)$ {\tt do}\\
08:      \indc  $N_d := \{ i | dist(i) = d \}$\\
09:      \indc  {\tt for each} $i \in N_d$ {\tt do}\\
10:          \indd  $x := \min_{j \in par(i)} \bar{y}_j$\\
11:          \indd  {\tt if} $(x < \hat{y}_i)$\\
12:             \inde  $\bar{y}_i := x$\\
13:          \indd  {\tt else}\\
14:             \inde  $\bar{y}_i := \hat{y}_i$\\
15:      \indc  {\tt end for}\\
16:    \indb {\tt end for}\\
{\tt end algorithm}\\
\noindent
{\tt Output}:\\
- $\bar{\by} = < \bar{y}_1, \bar{y}_2, \ldots, \bar{y}_{|V|}>$\\
\end{minipage}
}
\end{center}
\end{figure}

Fig.~\ref{fig-HTD} provides the pseudo code of the algorithm. 
Rows from $1$ to $4$ provide the distance of each node from the root, where distance means the maximum path length from the node to the root. To this end the classical Bellman-Ford algorithm~\cite{Cormen09} can be used: by recalling that if finds the shortest paths from a source node to all the other nodes of a weighted digraph, it is sufficient to invert the sign of each edge weight to obtain the maximum distance (longest path) from the root (rows $2$ and $3$). Note that $dist$ is a vector containing for each element $i$ the maximum distance of the node $i$ from the root.

The second block of the algorithm implements a per-level top-down visit of the graph (rows $5-16$).
Starting from the children of the root (level $1$) for each level of the graph the nodes are processed and the hierarchical top-down correction of the flat predictions $\hat{y}_i$, $i \in \{1, \ldots, |V| \}$ to the HTD-DAG ensemble prediction  $\bar{y}_i$ is performed according to eq.~\ref{eq:HTD}.

The first block $A$ (rows $1-4$) of the algorithm is dominated by the Bellman-Ford algorithm and has complexity $\mathcal{O}(|V|^2)$ for sparse graphs, while it is easy to see that the complexity of the second block $B$ (rows $5-16$) is linear in the number of vertices for sparse graphs.

\section{Hierarchical True Path Rule algorithm for DAGs (TPR-DAG)}
\label{sec:TPR}

This second algorithm represents a DAG extension of the {\em TPR} algorithm, originally proposed for tree-structured taxonomies~\cite{Vale09a,Vale11a}.
The main difference with respect to the original tree-version consists in the fact that the per-level traversal is now performed through two completely distinct steps: a bottom-up per level visit of the graph followed by a top-down visit, while in the original tree-version the per-level traversal is performed in an ``interleaved'' fashion (that is the bottom-up and top-down traversal are alternated at each level~\cite{Vale11a}). 
In the DAG version the separation of the bottom-up and top-down steps is necessary to assure the true path rule consistency of the predictions.

\begin{figure}[!h]
\begin{center}
\cornersize*{8pt}
\ovalbox{
\begin{minipage}{0.90\textwidth}
\caption{\bf{Hierarchical True Path Rule algorithm for DAGs (TPR-DAG)}}
\label{fig-TPR}

\noindent
{\tt Input}:\\
- $G = <V,E>$\\
- $V = \{1, 2, \ldots, |V| \}$\\
- $\hat{\by} = < \hat{y}_1, \hat{y}_2, \ldots, \hat{y}_{|V|}>, \quad  \hat{y}_i \in [0,1]$\\
- $\bt = < t_1, t_2, \ldots, t_{|V|}>, \quad  t_i \in [0,1]$\\
\noindent
{\tt begin algorithm}\\
01:  \inda   A. Compute $\forall i \in V$ the max distance from $root(G)$:\\
02:    \indb $E' := \{e' | e \in E, \, e' = - e \}$\\
03:    \indb $G' := < V, E' >$\\
04:    \indb $dist := $ Bellman.Ford$(G', root(G')$ \\
05:  \inda   B. Per-level bottom-up visit of $G$:\\ 
06:    \indb {\tt for each} $d$ {\tt from} $\max(dist)$  {\tt to} $1$ {\tt do}\\
07:      \indc  $N_d := \{ i | dist(i) = d \}$\\
08:      \indc  {\tt for each} $i \in N_d$ {\tt do}\\
09:          \indd  $\phi_i := \{ j \in child(i) | \bar{y}_j > t_j \}$ \\
10:          \indd  $\bar{y}_i := \frac{1}{1 + |\phi_i|} (\hat{y}_i + \sum_{j \in \phi_i} \bar{y}_j)$ \\
11:      \indc  {\tt end for}\\
12:    \indb {\tt end for}\\
13:  \inda   C. Per-level top-down visit of $G$:\\
14:    \indb $\bar{y}_{root(G)} := \hat{y}_{root(G)}$\\
15:    \indb {\tt for each} $d$ {\tt from} $1$ {\tt to} $\max(dist)$ {\tt do}\\
16:      \indc  $N_d := \{ i | dist(i) = d \}$\\
17:      \indc  {\tt for each} $i \in N_d$ {\tt do}\\
18:          \indd  $x := \min_{j \in par(i)} \bar{y}_j$\\
19:          \indd  {\tt if} $(x < \hat{y}_i)$\\
20:             \inde  $\bar{y}_i := x$\\
21:          \indd  {\tt else}\\
22:             \inde  $\bar{y}_i := \hat{y}_i$\\
23:      \indc  {\tt end for}\\
24:    \indb {\tt end for}\\
{\tt end algorithm}\\
\noindent
{\tt Output}:\\
- $\bar{\by} = < \bar{y}_1, \bar{y}_2, \ldots, \bar{y}_{|V|}>$\\
\end{minipage}
}
\end{center}
\end{figure}

The other main difference consists in the way the levels are computed: in this new DAG version the levels are constructed according to the maximum distance from the root, since this guarantees that in the top-down step all the ancestor nodes have been just processed, thus assuring the true path rule consistency of the predictions. 

Similarly to the tree-based version the {\em TPR} algorithm, the {\em TPR-DAG} adopts a per-level bottom-up traversal of the DAG, starting from the nodes most distant (in the sense of the maximum distance) from the root to correct the flat predictions $\hat{y}_i$:

\begin{equation} 
\bar{y}_i := \frac{1}{1 + |\phi_i|} (\hat{y}_i + \sum_{j \in \phi_i} \bar{y}_j)
\label{eq:TPR}
\end{equation}
where $\phi_i$ are the ``positive'' children of $i$.
\begin{equation} 
\phi_i := \{ j \in child(i) | \bar{y}_j > t_j \}
\label{eq:phi}
\end{equation}
The choice of the set of the ``positive'' children $\phi_i$, that is of the children responsible for the bottom-up propagation of the positive predictions, depends critically on the choice of the threshold $t_j$. Possible choices could be the following:
\begin{enumerate}[label=(\alph*)]
\item $t_j = \bar{t}, \quad \forall j \in V$: that is an equal threshold $\bar{t}$ is selected for all the nodes. For instance if the predictions represent probabilities it could be meaningful to a priori select $\bar{t} = 0.5$.
\item $t_j$ is selected to maximize the F-score estimated on the training data
\item $t_j$ is selected on the basis of the distribution of the positive examples available for the training set (e.g. $t_j$ could correspond to the $k^{th}$ percentile of the positive examples distribution).
\end{enumerate}
A different solution, that does not require an a priori or and experimentally selected threshold could consists in choosing those children that can increment the score of the node $i$ (that is  positive nodes are those that achieve a higher score than that of their parent):
\begin{equation} 
\phi_i := \{ j \in child(i) | \bar{y}_j > \hat{y}_i \}
\label{eq:phi2}
\end{equation}

Independently of the choice of the positive children, the following theorem holds:
\begin{theorem} Given a DAG $G = <V,E>$, a set of flat predictions $\hat{\by} = < \hat{y}_1, \hat{y}_2, \ldots, \hat{y}_{|V|}>$ for each class associated to each node $i \in \{1, \ldots, |V| \}$,  the {\em TPR-DAG} algorithm assures that $\forall i \in V, \; i \in anc(j) \Rightarrow \bar{y}_i \geq \bar{y}_i$.
\end{theorem}
The proof is omitted for brevity.

Fig.~\ref{fig-TPR} shows the high-level pseudo-code of the {\em TPR-DAG} algorithm.
The first four rows compute the maximum distance of each node from the root, using the Bellman-Ford algorithm.
The block $B$ (rows 5-12) performs a bottom-up visit of the graph and updates the predictions $\bar{y}_i$ of the {\em TPR-DAG} ensemble according to eq.~\ref{eq:TPR} and \ref{eq:phi}.
Note that this step propagates the ``positive'' predictions from bottom to top of the DAG, but 
does not assure the true path rule consistency of the predictions.
This is accomplished by the third block (rows $13-24$) that simply executes a hierarchical top-down step, in the same way of the {\em HTD-DAG} algorithm.

Also for the {\em TPR-DAG} algorithm the complexity is quadratic in the number of nodes for the block $A$, and it is easy to see that it is $\mathcal{O}(|V|)$ for both the $B$ and $C$ blocks when graphs are sparse.


\begin{figure}[!tb]
\begin{center}
\cornersize*{8pt}
\ovalbox{
\begin{minipage}{0.90\textwidth}
\caption{\bf{Hierarchical Isotonic True Path Rule algorithm for DAGs (ISO-TPR)}}
\label{fig-ISOTPR}

\noindent
{\tt Input}:\\
- $G = <V,E>$\\
- $V = \{1, 2, \ldots, |V| \}$\\
- $\hat{\by} = < \hat{y}_1, \hat{y}_2, \ldots, \hat{y}_{|V|}>, \quad  \hat{y}_i \in [0,1]$\\
- $\bt = < t_1, t_2, \ldots, t_{|V|}>, \quad  t_i \in [0,1]$\\
\noindent
{\tt begin algorithm}\\
01:  \inda   A. Compute $\forall i \in V$ the max distance from $root(G)$:\\
02:    \indb $E' := \{e' | e \in E, \, e' = - e \}$\\
03:    \indb $G' := < V, E' >$\\
04:    \indb $dist := $ Bellman.Ford$(G', root(G')$ \\
05:  \inda   B. Per-level bottom-up visit of $G$:\\ 
06:    \indb {\tt for each} $d$ {\tt from} $\max(dist)$  {\tt to} $1$ {\tt do}\\
07:      \indc  $N_d := \{ i | dist(i) = d \}$\\
08:      \indc  {\tt for each} $i \in N_d$ {\tt do}\\
09:          \indd  $\phi_i := \{ j \in child(i) | \bar{y}_j > t_j \}$ \\
10:          \indd  $\bar{y}_i := \frac{1}{1 + |\phi_i|} (\hat{y}_i + \sum_{j \in \phi_i} \bar{y}_j)$ \\
11:      \indc  {\tt end for}\\
12:    \indb {\tt end for}\\
13:  \inda   C. Isotonic correction:\\
14:    \indb $ \bar{\by} = \left\{
               \begin{array}{l}
                   \min_{\bar{\by}} \sum_{i \in V} (\hat{y}_i - \bar{y}_i )^2\\
                   \forall i, \quad  j \in par(i) \Rightarrow  \bar{y}_j  \geq \bar{y}_j
               \end{array}
       \right. 
      $\\
{\tt end algorithm}\\
\noindent
{\tt Output}:\\
- $\bar{\by} = < \bar{y}_1, \bar{y}_2, \ldots, \bar{y}_{|V|}>$\\
\end{minipage}
}
\end{center}
\end{figure}

A variant of the {\em TPR} algorithm for DAG is represented by the {\em ISO-TPR} algorithm (Fig.~\ref{fig-ISOTPR}). 
The main difference with respect to the {\em TPR-DAG} algorithm is constituted by the top-down step: instead of using the {\em HTD-DAG} top-down step a partial order isotonic regression approach is applied~\cite{Barlow72}. That is, the following optimization problem is solved (rows 13-14):
\begin{equation}
\bar{\by} = \left\{
               \begin{array}{l}
                   \min_{\bar{\by}} \sum_{i \in V} (\hat{y}_i - \bar{y}_i )^2\\
                   \forall i, \quad  j \in par(i) \Rightarrow  \bar{y}_j  \geq \bar{y}_j
               \end{array}
       \right. 
\label{eq:iso}
\end{equation}
In this way the true path rule constraints are maintained by construction, and the solution closest to flat predictions (in the sense of the squared error) that obeys the true path rule is selected.

It is worth noting that other variants similar to the weighted True Path Rule algorithms for tree-structured taxonomies~\cite{Vale09d,Vale12a} can be designed for DAGs, simply by substituting row $10$ of the {TPR-DAG} algorithm (Fig.~\ref{fig-TPR}). For instance we can obtain the  {\em TPR-w-DAG} algorithm by substituting row $10$ of the {TPR-DAG} algorithm with the following line of pseudocode:
\begin{equation}
\bar{y}_i := w \hat{y}_i + \frac{(1 - w)}{|\phi_i|} \sum_{j \in \phi_i} \bar{y}_j
\label{eq:tpr-w}
\end{equation}
In this approach a weight $w \in [0,1]$ is added to balance between the contribution of the node $i$ and that of its ``positive'' children.

As shown in~\cite{Vale11a}, the contribution of the descendants of a given node decays exponentially with their distance from the node itself.
To enhance the contribution of the most specific nodes to the overall decision of the ensemble a linear decaying or a constant contribution of the ``positive'' descendants could be considered instead:
\begin{equation}
\bar{y}_i := \frac{1}{1 + |\Delta_i|} ( \hat{y}_i + \sum_{j \in \Delta_i} \bar{y}_j )
\label{eq:tpr-desc}
\end{equation}
where 
\begin{equation}
\Delta_i = \{ j \in desc(i) | \bar{y}_j > t_j \}
\label{eq:delta}
\end{equation}
In this way all the ``positive'' descendants of node $i$ provide the same contribution to the ensemble prediction $\bar{y}_i$.
Analogously, we can design ``positive'' descendants whose contributions to $\bar{y}_i$ decays linearly with their distance from the root.

\section{Conclusions}
The {\em HTD-DAG} algorithm represents a simple and efficient approach to make consistent the predictions of flat classifiers in a DAG-structured taxonomy. 
The {\em TPR-DAG} and its variants including {\em ISO-TPR} represent an adaptation to DAGs of the previously proposed {\em TPR} algorithm for tree-structured taxonomies. {\em TPR-DAG} exploits the children (and recursively the descendants) of each node of the DAG-structured ensemble, to propagate from bottom to top the positive predictions of the classifiers associated to the most specific classes of the hierarchy. In this way positive predictions for the most specific classes are valued and passed to the less specific ones. A second top-down step re-establishes the true path rule constraints and propagate negative predictions towards the descendants. 

Experimental studies with DAG-structured taxonomies are planned to assess the effectiveness of the proposed methods for real-world hierarchical classification problems.

\bibliographystyle{plain}

\end{document}